\pdfoutput=1

\documentclass[11pt]{article}

\usepackage{acl}

\usepackage{times}
\usepackage{latexsym}

\usepackage[T1]{fontenc}

\usepackage[utf8]{inputenc}
\usepackage{times}
\usepackage{latexsym}
\usepackage{url}
\usepackage{enumitem}
\usepackage{amsmath,amssymb,amsfonts}
\usepackage{algorithmic}
\usepackage{caption}
\usepackage{subcaption}
\usepackage{graphicx}
\usepackage{textcomp}
\usepackage{url}
\usepackage{makecell}
\usepackage{multirow}
\usepackage{booktabs}
\usepackage{array}
\usepackage{microtype}
\usepackage{colortbl}

\usepackage{multirow}
\usepackage{booktabs}
\usepackage{hyperref}

\usepackage{hyperref}       
\usepackage{amsfonts}       
\usepackage{nicefrac}       
\usepackage{tcolorbox}
\usepackage{xcolor}[dvipsnames]

\newcommand{\thickhline}{%
    \noalign {\ifnum 0=`}\fi \hrule height 1pt
    \futurelet \reserved@a \@xhline
}

\usepackage{pifont}
\newcommand{\cmark}{O}%
\newcommand{\xmark}{\ding{55}}%
\definecolor{ForestGreen}{RGB}{34,139,34}
\usepackage{color}
\definecolor{lightgray}{gray}{0.9}

\newcommand{\inlinecode}[2]{\colorbox{lightgray}{\lstinline[language=#1]$#2$}}

\usepackage{comment}
\usepackage{todonotes}

\usepackage{microtype}
\usepackage[cjk]{kotex}
\newcommand\blfootnote[1]{%
  \begingroup
  \renewcommand\thefootnote{}\footnote{#1}%
  \addtocounter{footnote}{-1}%
  \endgroup
}

\definecolor{lp}{HTML}{CBC3E3}

\title{Evalverse: Unified and Accessible Library for Large Language Model Evaluation}

\author{Jihoo Kim, Wonho Song, Dahyun Kim, Yunsu Kim, Yungi Kim, Chanjun Park$^{\dagger}$\\
\\
  Upstage AI \\
  \texttt{\{jerry, ynot, kdahyun, yoonsoo, eddie, chanjun.park\}@upstage.ai}}

\begin{document}
\maketitle

\begin{abstract}
\blfootnote{$^\dagger$ Corresponding Author }
This paper introduces Evalverse\footnote{\url{https://github.com/UpstageAI/evalverse}}, a novel library that streamlines the evaluation of Large Language Models (LLMs) by unifying disparate evaluation tools into a single, user-friendly framework. Evalverse enables individuals with limited knowledge of artificial intelligence to easily request LLM evaluations and receive detailed reports, facilitated by an integration with communication platforms like Slack. Thus, Evalverse serves as a powerful tool for the comprehensive assessment of LLMs, offering both researchers and practitioners a centralized and easily accessible evaluation framework. Finally, we also provide a demo video for Evalverse, showcasing its capabilities and implementation in a two-minute format\footnote{\url{https://www.youtube.com/watch?v=-VviAutjpgM}}.
\end{abstract}

\section{Introduction}
In recent years, the rapid advancements in Large Language Models (LLMs) have significantly transformed the computational linguistics landscape, presenting novel opportunities and challenges~\cite{wei2022emergent,zhao2023survey}. Due to the vast scale and complexity of LLMs~\cite{kaplan2020scaling}, they have demonstrated remarkable capabilities across numerous applications~\cite{hadi2023survey}, ranging from natural language understanding and generation to more specialized tasks such as summarization~\cite{jin2024comprehensive}, translation~\cite{hendy2023good}, and question-answering~\cite{zhuang2024toolqa}. However, the sheer pace of LLM development has led to a fragmented ecosystem of evaluation tools and methodologies~\cite{chang2023survey,guo2023evaluating}. This fragmentation not only hinders the comparative assessment of LLMs, but also places a considerable barrier to entry for both researchers and practitioners.

\begin{figure}
    \centering
    \includegraphics[width=0.42\textwidth]{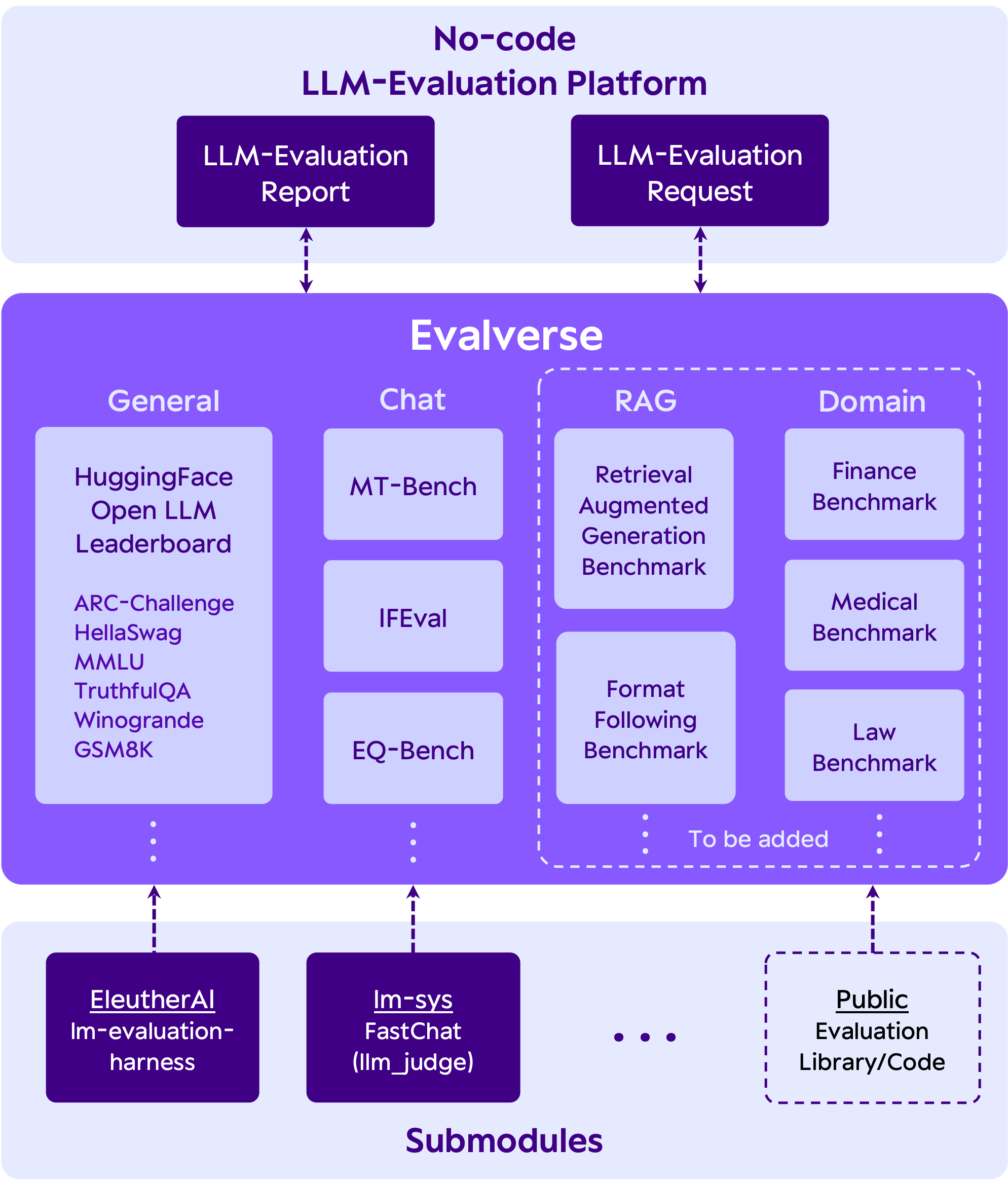}
    \caption{Overview of Evalverse. Users can interact with Evalverse in a no-code manner. External benchmark frameworks are integrated as submodules.}
    \label{fig:evalverse_overview}
\end{figure}

Recognizing the critical need for a more unified and accessible framework for LLM evaluation, we introduce Evalverse with the overview depicted in Figure~\ref{fig:evalverse_overview} -- a novel library that centralizes various evaluation methodologies. Evalverse built such that it can function as a unified and expandable library for LLM evaluation while also lowering the technical barrier to entry of LLM evaluation. 

To achieve the former, we integrate existing evaluation frameworks, such as lm-evaluation-harness~\cite{eval-harness} and FastChat~\cite{zheng2024judging}, as submodules, allowing an easy extension of new benchmarks.
These added submodules can reflect recent changes, allowing Evalverse to remain up-to-date with relative ease.
On the other hand, we also implement no-code evaluation features that utilize communication platforms such as Slack\footnote{\url{https://slack.com/}}, making LLM evaluation more accesible for individuals with less programming proficiency.

\begin{figure*}[t!]
    \centering
    \includegraphics[width=0.85\textwidth]{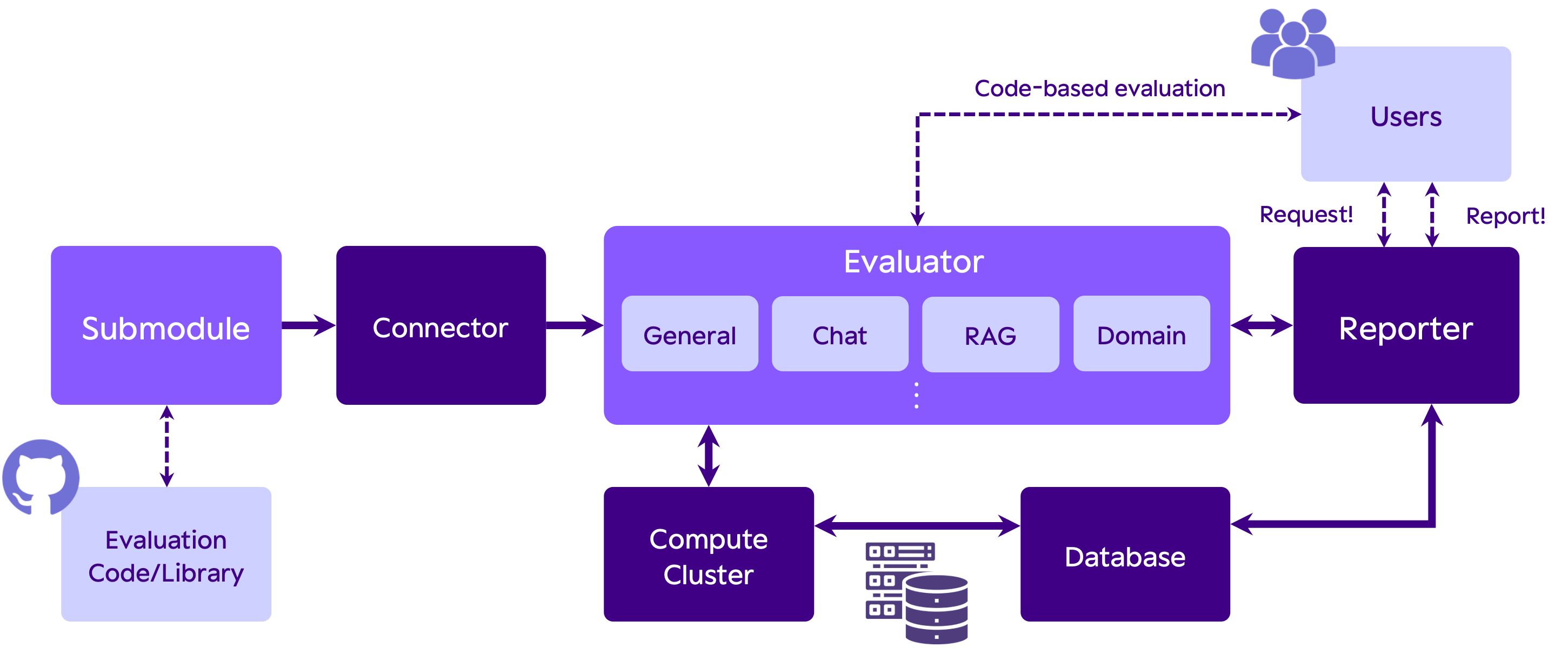}
    \caption{The system architecture of Evalverse. Users can use the Evaluator directly for code-based evaluation, or interact with the Reporter for a no-code approach to LLM evaluation.}
    \label{fig:evalverse_architecture}
\end{figure*}

This paper provides an in-depth examination of the architecture and functionality of Evalverse, illustrating how it addresses the current challenges in LLM evaluation. Some of the key features of Evalverse include no-code evaluation and a unified and expandable library for LLM benchmarks, enhancing the efficiency and accessibility.

\begin{table*}[]
\resizebox{1.0\linewidth}{!}{
\begin{tabular}{lccccccccccc}
\toprule
                          & \textbf{General} & \multicolumn{3}{c}{\textbf{Chat}} & \textbf{RAG} & \multicolumn{3}{c}{\textbf{Domain}} & \multicolumn{3}{c}{\textbf{Additional Features}}  \\ \cmidrule(lr){1-1}\cmidrule(lr){2-2}\cmidrule(lr){3-5}\cmidrule(lr){6-6}\cmidrule(lr){7-9}\cmidrule(lr){10-12}
\textbf{Evaluation Framework}      & H6 Avg   & MT-Bench   & IFEval   & EQ-Bench  & RGB         & Finance     & Medical     & Law     & Leaderbaord & Eval Report & No-Code Eval \\ \cmidrule(lr){1-1}\cmidrule(lr){2-2}\cmidrule(lr){3-5}\cmidrule(lr){6-6}\cmidrule(lr){7-9}\cmidrule(lr){10-12}
lm-evaluation-harness     & {\color{ForestGreen} \cmark}        & {\color{red} \xmark}          & {\color{ForestGreen} \cmark}        & {\color{ForestGreen} \cmark}         & {\color{red} \xmark}           & {\color{red} \xmark}           & {\color{ForestGreen} \cmark}           & {\color{red} \xmark}       & {\color{red} \xmark}           & {\color{red} \xmark}           & {\color{red} \xmark}            \\
FastChat                  & {\color{red} \xmark}        & {\color{ForestGreen} \cmark}          & {\color{red} \xmark}        & {\color{red} \xmark}         & {\color{red} \xmark}           & {\color{red} \xmark}           & {\color{red} \xmark}           & {\color{red} \xmark}       & {\color{ForestGreen} \cmark}           & {\color{red} \xmark}           & {\color{red} \xmark}            \\
OpenCompass               & {\color{ForestGreen} \cmark}        & {\color{ForestGreen} \cmark}          & {\color{ForestGreen} \cmark}        & {\color{red} \xmark}         & {\color{red} \xmark}           & {\color{ForestGreen} \cmark}           & {\color{ForestGreen} \cmark}           & {\color{ForestGreen} \cmark}       & {\color{ForestGreen} \cmark}           & {\color{red} \xmark}           & {\color{ForestGreen} \cmark}            \\
LightEval                 & {\color{ForestGreen} \cmark}        & {\color{red} \xmark}          & {\color{ForestGreen} \cmark}        & {\color{red} \xmark}         & {\color{red} \xmark}           & {\color{red} \xmark}           & {\color{ForestGreen} \cmark}           & {\color{ForestGreen} \cmark}       & {\color{ForestGreen} \cmark}           & {\color{red} \xmark}           & {\color{red} \xmark}            \\
\cellcolor{lp!60}\textbf{Evalverse (Ours)} & \cellcolor{lp!60}{\color{ForestGreen} \cmark}        & \cellcolor{lp!60}{\color{ForestGreen} \cmark}          & \cellcolor{lp!60}{\color{ForestGreen} \cmark}        & \cellcolor{lp!60}{\color{ForestGreen} \cmark}         & \cellcolor{lp!60}{\color{blue} $\triangle$}           & \cellcolor{lp!60}{\color{blue} $\triangle$}           & \cellcolor{lp!60}{\color{blue} $\triangle$}           & \cellcolor{lp!60}{\color{blue} $\triangle$}       & \cellcolor{lp!60}{\color{red} \xmark}           & \cellcolor{lp!60}{\color{ForestGreen} \cmark}           & \cellcolor{lp!60}{\color{ForestGreen} \cmark}            \\ \bottomrule
\end{tabular}
}
\caption{Comparison between LLM evaluation frameworks. Note that Evalverse incorporates all of the shown benchmarks in for ``General'' and ``Chat'' evaluation, respectively. Further, we are actively expanding Evalverse to include benchmarks for RAG and other domain specific evaluations as well, indicated by the blue triangle. Further, Evalverse supports no-code evaluation and reports, unlike other frameworks.}
\label{tab:eval_frameworks}
\end{table*}

\section{Related Work and Background}\label{sec:related_work}
\subsection{LLM Evaluation Aspects}
There are multiple aspects of LLM evaluation, which can be divided into the following four categories: i) general performance; ii) performance for chat applications; iii) performance for Retrieval Augmented Generation (RAG)~\cite{lewis2020retrieval}; iv) performance for various domains.

\paragraph{General performance.}
The Hugging Face Open LLM Leaderboard~\cite{beeching2023open} is primarily utilized for evaluation general performance. The leaderboard uses a total of six benchmarks, AI2 Reasoning Challenge~\cite{clark2018think}, HellaSwag~\cite{zellers2019hellaswag}, Massive Multitask Language Understanding (MMLU)~\cite{hendrycks2020measuring}, TruthfulQA~\cite{lin2021truthfulqa}, Winogrande~\cite{sakaguchi2021winogrande}, and GSM8k~\cite{cobbe2021training}, and the average of these scores is commonly referred to as H6 Avg.

\paragraph{Performance for chat applications.} One of the primary use cases for LLMs is chat applications~\cite{team2023gemini,achiam2023gpt}. It is crucial to measure whether LLMs follow the user's instructions properly and work effectively in a multi-turn environment. The representative methods for evaluating these chat abilities are MT-Bench~\cite{zheng2024judging}, IFEval~\cite{zhou2023instruction}, and EQ-Bench~\cite{paech2023eq}.

\paragraph{Performance for RAG.} Pre-trained or fine-tuned LLMs alone may not be sufficient to meet business-level requirements. Therefore, RAG can be an appropriate solution, which involves retrieving documents related to the user queries and providing them as input context to the LLMs. To judge the performance of the LLMs in terms of RAG performance, \citet{chen2023benchmarking} introduces Retrieval-Augmented Generation Benchmark (RGB). Further, \citet{xia2024fofo} presents Format-Following benchmark (FoFo) for evaluating the ability to follow specific formats, which is important for more complex RAG applications as they heavily depend on the intermediate outputs adhering to pre-defined structures.

\paragraph{Performance for various domains.} There are many applications of LLMs in various domains such as finance, healthcare, and law. The FinGPT Benchmark~\cite{wang2023fingpt}, MultiMedQA~\cite{singhal2023large}, and LegalBench~\cite{guha2022legalbench} correspond to the financial, medical, and legal domain, respectively.

\subsection{LLM Evaluation Frameworks}
There exists other evaluation frameworks for measuring the performance of LLMs across multiple benchmarks. Eleuther AI's lm-evaluation-harness~\cite{eval-harness} is a widely used framework, where over 60 tasks are supported such as H6 Avg, IFEval, and EQ-Bench. LMSYS Org's FastChat~\cite{zheng2024judging} supports LLM-Judge to evaluate MT-Bench. OpenCompass\footnote{\url{https://github.com/open-compass/OpenCompass/}} is an LLM evaluation platform supporting evaluations not only for H6 Avg, MT-Bench and IFEval but also for multiple domains like Finance, Healthcare, and Law. The most recently released LightEval\footnote{\url{https://github.com/huggingface/lighteval}} by HuggingFace is built on top of EleutherAI's lm-evaluation harness. The difference between these frameworks and Evalverse is summarized in Table~\ref{tab:eval_frameworks}.

\section{Evalverse}

\subsection{Why Evalverse?}
The core motivation behind Evalverse is to facilitate a unified and expandable library for LLM evaluation, while also being more easily accessible than other existing frameworks.
To that end, we integrate benchmarks in a way that is less burdensome to keep them up-to-date.
Further, we engineer a {\it no-code approach} for LLM evaluation, thereby broadening the user base beyond those with coding proficiency.
This sets Evalverse apart from conventional evaluation frameworks~\cite{resnik2010evaluation} that often necessitate programming skills.

This paper elucidates the architecture and functional capabilities. We posit that the design principles adopted in Evalverse could serve as a blueprint for other evaluation frameworks as well.

\subsection{Evalverse Architecture}
We explain the system architecture of Evalverse to facilitate a unified evaluation framework whilst also supporting no-code evaluation.
Evalverse consists of the following six primary components: Submodule, Connector, Evaluator, Compute Cluster, Database, and Reporter.
The overall architecture of Evalverse is illustrated in Figure~\ref{fig:evalverse_architecture}.

\paragraph{Submodule.} The Submodule serves as the evaluation engine that is responsible for the heavy lifting involved in evaluating LLMs. Publicly available LLM evaluation libraries can be integrated into Evalverse as submodules. This component makes Evalverse expandable, thereby ensuring that the library remains up-to-date.

\paragraph{Connector.} The Connector plays a role in linking the Submodules with the Evaluator. It contains evaluation scripts, along with the necessary arguments, from various external libraries.

\paragraph{Evaluator.} The Evaluator performs the requested evaluations on the Compute Cluster by utilizing the evaluation scripts from the Connector.
The Evaluator can receive evaluation requests either from the Reporter, which facilitates a no-code evaluation approach, or directly from the end-user for code-based evaluation.

\paragraph{Compute Cluster.}
The Compute Cluster is the collection of hardware accelerators needed to execute the LLM evaluation processes.
When the Evaluator schedules an evaluation job to be ran, the Compute Cluster fetches the required model and data files from the Database.
The results of the evaluation jobs are sent to the Database for storage.

\paragraph{Database.} 
The Database stores the model files and data needed in the evaluation processes, along with evaluation results.
The stored evaluation results are used by the Reporter to create evaluation reports for the user.

\paragraph{Reporter.} The Reporter handles the evaluation and report requests sent by the users, allowing for a no-code approach to LLM evaluation. 
The Reporter sends the requested evaluation jobs to the Evaluator and fetches the evaluation results from the Database, which are sent to the user via an external communication platform such as Slack.
Through this, users can receive table and figure that summarize evaluation results.

\subsection{Evalverse Functionality}
We detail the no-code, unified, and expandable evaluation as core functionalities of Evalverse, derived from its system architecture.

\paragraph{No-code evaluation.}
Evalverse supports no-code evaluation using the Reporter explained in the previous section.
We have chosen Slack as the initial external communication tool for the no-code evaluation feature, owing to its popular use among numerous companies and communities alike.\footnote{Expansion to other communication tools are set as important milestones in the development road-map.} 
A detailed example usage of no-code evaluation is given in Section~\ref{sec:no-code-eval}.

Further, Evalverse also supports a no-code evaluation report feature, where average scores and rankings for just the selected models are retrieved from the Database.
This functionality allows non-technical personnel to proactively retrieve evaluation results without having to ask someone with more programming proficiency. 
Example usage is illustrated in Section~\ref{sec:no-code-look-up}.

\paragraph{Unified and expandable evaluation.}
For unified and expandable evaluation, Evalverse utilizes Git submodules\footnote{\url{https://git-scm.com/book/en/v2/Git-Tools-Submodules}} to integrate external evaluation frameworks such as lm-evaluation-harness~\cite{eval-harness} and FastChat~\cite{zheng2024judging}. 
Thus, one can easily add new submodules to support more external evaluation frameworks. Not only that, one can always fetch upstream changes of the submodules to stay up-to-date with evaluation processes in the fast-paced LLM field.

Evalverse includes IFEval~\cite{zhou2023instruction} and EQ-Bench~\cite{paech2023eq} which are designed for more nuanced evaluation of LLMs for chat applications.
Furthermore, RGB~\cite{chen2023benchmarking}, FoFo~\cite{xia2024fofo}, FinGPT~\cite{wang2023fingpt}, MultiMedQA~\cite{liu2024evaluation} and LegalBench~\cite{guha2022legalbench} are being added to expand the evaluation suite to RAG, finance, medical, and legal capabilities, respectively.

The unified nature of Evalverse allows a one-step installation of all the required dependencies for different LLM evaluations.
Further, one can aggregate and manage common arguments across multiple benchmarks, such as model name or path.

\subsection{Evalverse Tour}
We demonstrate how to use Evalverse from installation to executing no-code and code-based evaluation processes.

\paragraph{Installation.}
One can clone the Evalverse repository and install all the necessary packages at once with the following command:
\begin{lstlisting}[language=bash]
# Evalverse and submodules
git clone --recursive https://github.com/UpstageAI/evalverse

# Install the required packages
cd evalverse
pip install -e .
\end{lstlisting}
Unlike a typical \inlinecode{bash}{git clone}, the additional \inlinecode{bash}{--recursive} option ensures that the submodules are also cloned.

\paragraph{Configuration.}
We recommend using a ``.env'' file to configure the required environment variables ({\it e.g.}, API keys), similar to the following example:
\begin{lstlisting}[language=bash]
# .env
OPENAI_API_KEY=sk-...

SLACK_BOT_TOKEN=xoxb-...
SLACK_APP_TOKEN=xapp-...
\end{lstlisting}
The ``OpenAI\_API\_Key'' is used to call the GPT-4 API~\cite{openai2023gpt4} in LLM-as-judge evaluation methods such as the MT-bench implemented in FastChat~\cite{zheng2024judging}. The ``Slack\_BOT\_Token'' and ``Slack\_APP\_Token'' are needed for the no-code evaluation feature via Slack, implemented in the Reporter.

\begin{figure}[t!]
    \centering
    \includegraphics[width=0.48\textwidth]{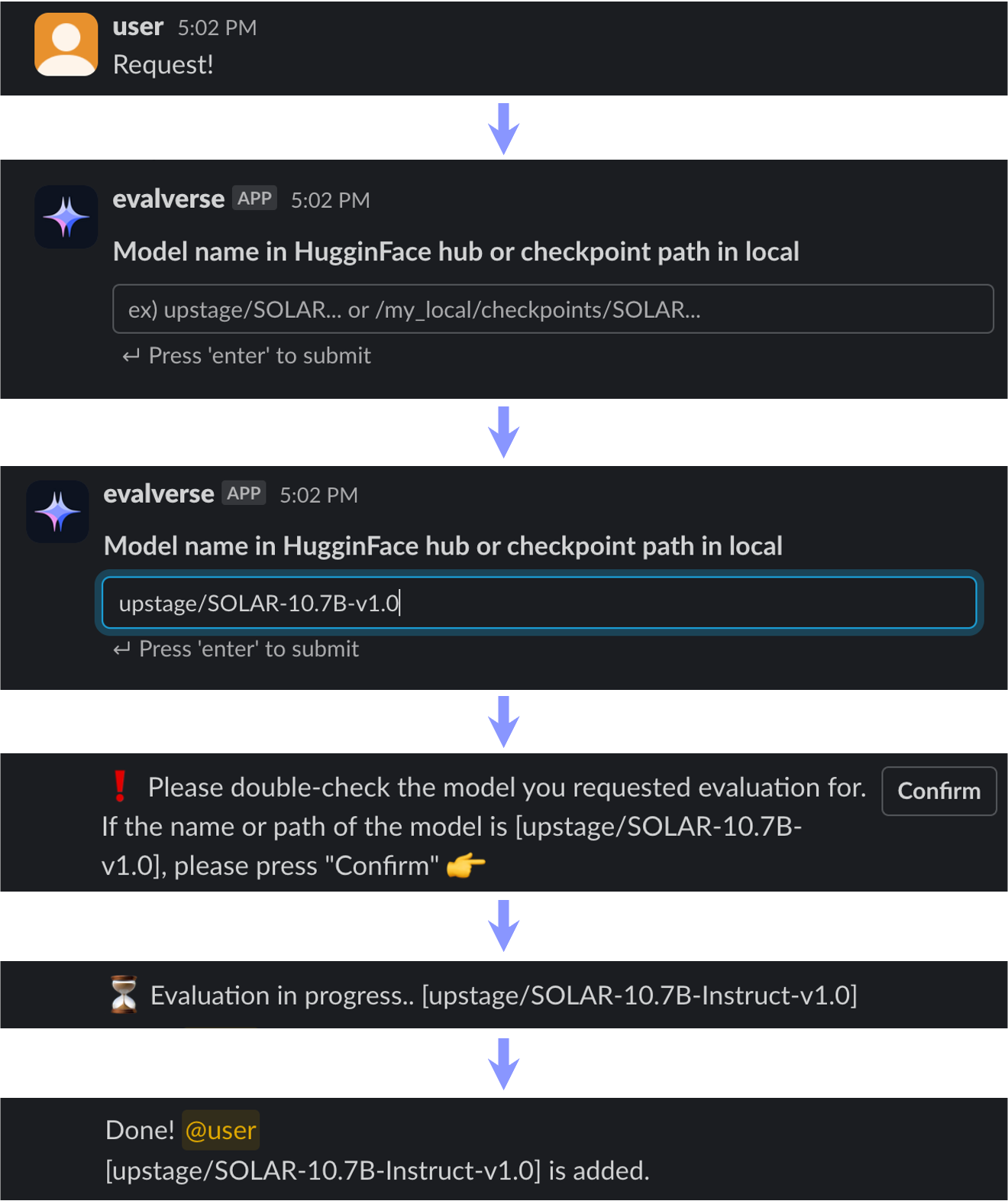}
    \caption{No-code evaluation request with Slack bot.}
    \label{fig:evalverse_request}
\end{figure}

\paragraph{No-code evaluation.}
\label{sec:no-code-eval}
Evalverse supports no-code evaluation via Slack requests, as depicted in Figure~\ref{fig:evalverse_request}.
The user types ``Request!'' in a direct message or Slack channel with an activate Evalverse Slack bot.
The Slack bot asks the user to enter the model name in the Huggingface hub~\cite{wolf2019huggingface} or the local model directory path.
Then, the Slack bot asks the user for confirmation and then launches an evaluation job on the remote Compute Cluster.
The Compute Cluster fetches the model file and necessary benchmark data caches (if present) from the Database and executes the evaluation process.
After the evaluation job is finished, an indication is sent to the user.
During the entire process, the user only interacts with the Slack bot with no programming involved.

\begin{figure}[t!]
    \centering
    \includegraphics[width=0.39\textwidth]{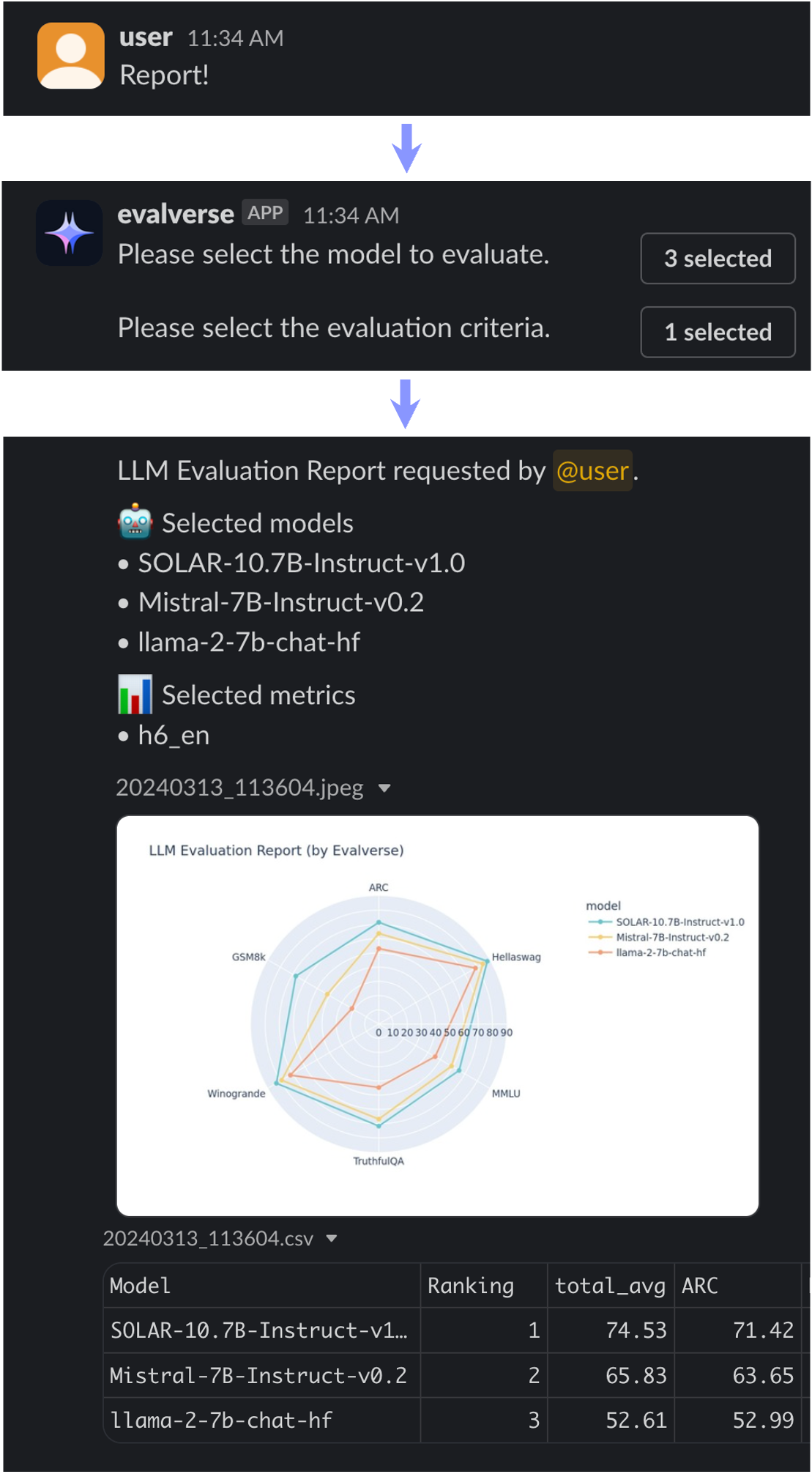}
    \caption{No-code evaluation report with Slack bot.}
    \label{fig:evalverse_report}
\end{figure}

\paragraph{No-code evaluation results look-up.}
\label{sec:no-code-look-up}
In addition to requesting new evaluations, Evalverse can also provide evaluation reports on finished evaluation in a no-code manner.
To receive the evaluation report, the user first types ``Report!'', similar to the evaluation request. Then, the Slack bot will ask the user to select the models and evaluation criteria.
For the selected model and evaluation criteria, Evalverse calculates the average scores and rankings using the evaluation results stored in the Database and provides a report with a performance table and a visualized graph as illustrated in Figure~\ref{fig:evalverse_report}.

\begin{table}
\centering
\resizebox{1.0\linewidth}{!}{
\begin{tabular}{ccc|cc}
\toprule
Engine & \# Few-shots & Dtype & SOLAR-10.7B-v1.0 & Mistral-7B-v0.1 \\
\midrule
hf & 5 & float16 & 64.38 & 62.59 \\ 
vllm & 5 & float16 & 64.36 & 62.65 \\
hf & 1 & float16 & 62.54 & 60.56 \\ 
hf & 5 & int8 & 64.24 & 62.51 \\
\bottomrule
\end{tabular}
}
\caption{MMLU scores depending on different inference engine options such as ``hf'', HuggingFace transformers~\cite{jain2022hugging}, or ``vllm'', the vLLM framework~\cite{kwon2023efficient}, and other options such as the data type (``dtype'') and number of few-shots.}
\label{tab:MMLU_Score_Difference_By_Evaluation_Settings}
\end{table}

\begin{table*}[t!]
\centering
\resizebox{0.7\linewidth}{!}{
\begin{tabular}{lcccccccccccc}
\toprule
\multirow{2}{*}{Model} & \multicolumn{2}{c}{H6} & \multicolumn{2}{c}{MT-Bench} & \multicolumn{2}{c}{EQ-Bench} & \multicolumn{2}{c}{IFEval} \\ 
\cmidrule(lr){2-9}
& orig & evalverse & orig & evalverse & orig & evalverse & orig & evalverse \\
\midrule
SOLAR 10.7B Instruct & 74.53 & 74.53 & 7.569 & 7.580 & 72.31 & 73.34 & - & 0.5370 \\
Mistral 7B Instruct & 65.82 & 65.82 & 7.466 & 7.600 & 70.05 & 66.57 & - & 0.5823 \\
Llama 2 7B Chat & 52.61 & 52.61 & 6.541 & 6.509 &35.09 & 37.76 & - & 0.4325 \\ 
Qwen 1.5 7B Chat & 55.66 & 55.66 & 7.606 & 7.575 & 57.33 & 51.33 & - & 0.4797 \\ 
\bottomrule
\end{tabular}
}
\caption{Comparison of evaluation results between the original (orig) repository and Evalverse for H6, MT-Bench, and EQ-Bench. The results show small differences compared to the original for benchmarks with no intentional modifications (H6, MT-Bench). The difference in EQ-Bench is mostly due to an intended modification of the prompts used in evaluation.}
\label{tab:tab2_1}
\end{table*}

\begin{table}[t!]
\centering
\resizebox{0.75\linewidth}{!}{
\begin{tabular}{ccccc}
\toprule
\multirow{2}{*}{Tools} & \multicolumn{4}{c}{Evaluation Time} \\
& H6 & MT-Bench & EQ-Bench & IFEval \\
\midrule
Original repo & 32.3 & 7.6 & 11.2 & - \\ 
Evalverse & 31.2 & 7.5 & 5.6 & 2.45 \\ 
\bottomrule
\end{tabular}
}
\caption{Evaluation time differences between the original repository and Evalverse for the Solar 10.7B Instruct model. Time units are expressed in minutes.}
\label{tab:tab3}
\end{table}

\paragraph{Code-based evaluation.}
In addition to the no-code evaluation features, one can conduct code-based evaluations for a more fine-grained control.
Evalverse supports running multiple benchmarks with a single Python script as detailed below.
\begin{lstlisting}[language=bash]
python3 evaluator.py \
    --ckpt_path {model_path} \
    --{benchmark_A} \
    --{benchmark_B} \
    --{args}
\end{lstlisting}
The \inlinecode{bash}{--ckpt-path} is a common argument used in all benchmarks, where the model name from the Hugging Face Hub or the local path of the model is given.
To evaluate a specific set of benchmarks, one can do so by adding the corresponding argument.
For a concrete example, the \inlinecode{bash}{--h6_en} argument is used for the H6 benchmark in the Open LLM Leaderboard~\cite{beeching2023open} implemented with lm-evaluation-harness, and the \inlinecode{bash}{--mt_bench} argument is used for MT-Bench implemented with FastChat.
Then, using 8 GPUs for data parallelism, one can perform evaluation on the aforementioned two benchmarks with the following command:
\begin{lstlisting}[language=bash]
python3 evaluator.py \
    --ckpt_path upstage/SOLAR-10.7B-Instruct-v1.0 \
    --h6_en \
    --mt_bench \
    --data_parallel 8
\end{lstlisting}

\section{Evaluation Comparisons}

We compare the evaluation results using Evalverse and the original implementation whenever possible.
The evaluated models include various open-source models such as Llama 2~\cite{touvron2023llama}, Mistral~\cite{jiang2023mistral}, Qwen 1.5~\cite{bai2023qwen}, and SOLAR~\cite{kim2023solar}.

\paragraph{Differences from the original implementation.}
When creating Evalverse, we adopted external frameworks as submodules, sometimes with intentional modifications.
First, the EQ-Bench in Evalverse uses the prompt in the original release of EQ-Bench version 2, whereas the upstream original repository uses the prompt in version 2.2. Version 2 uses revision prompts where it asks the model to revise its own answers if needed. In contrast, the prompts in version 2.2 do not use such revision prompts.
Once the changes in the upstream codebase are stabilized, the Evalverse submodule will be subsequently updated.

Further, the H6 benchmark implemented in lm-eval-harness supports a wide range of evaluation options, some of which may affect the evaluation results as shown in Table~\ref{tab:MMLU_Score_Difference_By_Evaluation_Settings}.
The table shows that the difference in the engine, dtype, and number of few-shot options can easily change the benchmark scores.
Thus, in the H6 benchmark of Evalverse, we fix the number of few-shots for to those used in the Open LLM Leaderboard and use the ``hf'' engine and ``float16'' dtype exclusively.

\paragraph{Reproducibility.}
To ensure that the benchmark scores from the original repositories are reproducible with Evalverse, we evaluate various open source models using the original implementation and Evalverse and summarize the results in Table~\ref{tab:tab2_1}.

The table shows that benchmarks with little modification (H6, MT-Bench) produce same or almost same scores as the original implementation, as the evaluation is done by using the submodules that are the no or little modifications from the original implementation. We also note that the score differences in MT-Bench are from the randomness of using LLM-as-a-judge.
On the other hand, the EQ-Bench benchmark results in a relatively larger score gap when compared to the original, due to the aforementioned intended modifications.
We could not compare to the original IFEval, since its implementation contains only the core logic and data, without any evaluation scripts.

\paragraph{Evaluation speed.}
We also compare evaluation speed of using Evalverse with that of the original implementation in Table~\ref{tab:tab3}.
The evaluation time with Evalverse and the original implementation for the H6, MT-Bench, EQ-Bench, and IFEval benchmarks using the SOLAR 10.7B Instruct model with 8$\times$A100 GPUs.
The H6 and MT-Bench have little evaluation time differences, whereas EQ-Bench evaluation time using Evalverse is faster for Evalverse.
The main reason is the added data parallelism support in the Evalverse submodule.

\paragraph{Evaluation of Open Source Models}\label{sec:evaluation}
\begin{table*}[]
\centering
\resizebox{0.95\linewidth}{!}{
\begin{tabular}{lccccccccc}
\toprule
Model & ARC & HellaSwag & MMLU & TruthfulQA & Winogrande & GSM8K & MT-Bench & EQ-Bench & IFEval \\
\midrule
 & \multicolumn{9}{c}{\textit{Pre-trained Models}} \\
\midrule
Mistral 7B & 61.43 & 83.31 & 62.64 & 42.62 & 79.16 & 37.83 & - & - & - \\ 
Solar 10.7B & 61.77 & 84.52 & 64.16 & 45.65 & 83.19 & 57.24 & - & - & - \\ 
Yi 34B & 65.44 & 85.75 & 76.51 & 56.27 & 83.19 & 65.73 & - & - & - \\ 
Mixtral 8x7B & 67.41 & 86.65 & 70.31 & 48.52 & 82.32 & 57.85 & - & - & - \\ 
Llama 2 70B & 67.58 & 87.00 & 68.83 & 44.81 & 83.35 & 52.62 & - & - & - \\ 
Qwen 1.5 72B & 66.21 & 85.97 & 77.25 & 59.57 & 82.72 & 68.69 & - & - & - \\ 
\midrule
 & \multicolumn{9}{c}{\textit{Fine-tuned Models}} \\
\midrule
Mistral 7B Instruct & 63.65 & 84.63 & 59.10 & 66.81 & 78.93 & 41.85 & 7.600 & 66.57 & 0.5823 \\ 
Solar 10.7B Instruct & 71.42 & 88.20 & 65.28 & 71.71 & 83.19 & 67.40 & 7.580 & 73.34 & 0.5370 \\ 
Yi 34B Chat & 65.18 & 84.28 & 74.98 & 55.40 & 80.35 & 34.50 & 7.641 & 72.35 & 0.3577 \\ 
Mixtral 8x7B Instruct & 70.39 & 87.31 & 70.30 & 63.34 & 82.00 & 64.97 & 8.200 & 72.97 & 0.5850 \\ 
Llama 2 70B Chat & 65.36 & 85.72 & 62.70 & 53.09 & 79.72 & 52.84 & 7.142 & 70.14 & 0.5370 \\ 
Qwen 1.5 72B Chat & 67.58 & 86.28 & 77.70 & 63.11 & 79.72 & 29.11 & 8.347 & 82.81 & 0.6146 \\ 

\bottomrule
\end{tabular}
}
\caption{Evaluation of open source models on various benchmarks using Evalverse.}
\label{tab:tab1}
\end{table*}

In Table~\ref{tab:tab1}, multiple open source models are evaluated using Evalverse for H6, MT-Bench, EQ-Bench, and IFEval benchmarks, respectively. The evaluated models are divided into two categories of pre-trained and fine-tuned models. 
For pre-trained models, we measured H6 scores to assess the the base reasoning and knowledge capabilities of the models, while fine-tuned models were additionally evaluated on MT-Bench, EQ-Bench, and IFEval benchmarks to assess their multi-turn chat and instruction following ability.
We used 8$\times$A100 GPUs for evaluation, along with 8-bit quantization for larger models such as Mixtral 8$\times$7B and Llama 2 70B.

\section{Conclusion}
We introduce Evalverse, a unified library for LLM evaluation that is easily expandable and accessible through no-code evaluation features. 
External benchmarks can be added via submodules, which makes addition of new benchmarks relatively easy while also making it possible for the added submodules to integrate upstream changes that may occur.
Using communication platforms such as Slack, users can request evaluation jobs and query evaluation results via Slack messages, enabling a no-code LLM evaluations.
We hope that by open-sourcing Evalverse, LLM evaluation can become more accessible and centralized, fueling further LLM development.

\section*{Acknowledgments}
This work was supported by Institute of Information \& Communications Technology Planning \& Evaluation(IITP) grant funded by the Korea government(MSIT) (No. RS-2024-00338140, Development of learning and utilization technology to reflect sustainability of generative language models and up-to-dateness over time).

\section*{Limitations}
While Evalverse represents a significant step forward in the evaluation of Large Language Models (LLMs), there are inherent limitations to our approach. First, as the landscape of LLM evaluation is rapidly evolving, keeping Evalverse up-to-date with the latest tools and methodologies poses an ongoing commitment despite our best efforts to make the update process relatively easy. Second, while we aim to make the evaluation accessible via the no-code features in Evalverse, accurately interpreting the results may still require specialized knowledge. Additionally, our reliance on community contributions to expand and update the library could lead to disparities in the coverage of evaluation tools, potentially affecting the comprehensiveness of Evalverse. Lastly, while integrating with platforms like Slack enhances accessibility, it also introduces dependencies on third-party services, which may affect the long-term sustainability and adaptability of Evalverse.

\section*{Ethics Statement}
In our Ethics Statement, we highlight the commitment of Evalverse to uphold ethical standards in the evaluation of Large Language Models (LLMs). We acknowledge the potential ethical issues, including privacy, security, and bias, associated with LLM evaluation. Evalverse is designed with a focus on transparency, accountability, and fairness, aiming to mitigate these concerns by promoting ethical research practices. This includes careful consideration of data sources, the impact on diverse communities, and efforts to reduce bias.

We stress the importance of responsible LLM use, advocating for evaluations that respect user privacy and data security. Evalverse is intended to foster an inclusive community of researchers by providing accessible evaluation tools and encouraging contributions from a broad spectrum of individuals. This approach not only addresses ethical concerns but also enhances the quality and inclusivity of LLM research. Our Ethics Statement reflects our dedication to advancing computational linguistics ethically, ensuring that LLM innovations consider their wider social and ethical impact.

\bibliography{anthology,custom}
\bibliographystyle{acl_natbib}

\end{document}